Paper accepted at the European Conf. on Computer Vision (ECCV), Munich, 2018.

# Asynchronous, Photometric Feature Tracking using Events and Frames

Daniel Gehrig, Henri Rebecq, Guillermo Gallego, and Davide Scaramuzza

Depts. Informatics and Neuroinformatics
University of Zurich and ETH Zurich

**Abstract.** We present a method that leverages the complementarity of event cameras and standard cameras to track visual features with low-latency. Event cameras are novel sensors that output pixel-level brightness changes, called "events". They offer significant advantages over standard cameras, namely a very high dynamic range, no motion blur, and a latency in the order of microseconds. However, because the same scene pattern can produce different events depending on the motion direction, establishing event correspondences across time is challenging. By contrast, standard cameras provide intensity measurements (frames) that do not depend on motion direction. Our method extracts features on frames and subsequently tracks them asynchronously using events, thereby exploiting the best of both types of data: the frames provide a photometric representation that does not depend on motion direction and the events provide low-latency updates. In contrast to previous works, which are based on heuristics, this is the first principled method that uses raw intensity measurements directly, based on a generative event model within a maximum-likelihood framework. As a result, our method produces feature tracks that are both more accurate (subpixel accuracy) and longer than the state of the art, across a wide variety of scenes.

## Multimedia Material

A supplemental video for this work is available at https://youtu.be/A7UfeUnG6c4

## 1 Introduction

Event cameras, such as the Dynamic Vision Sensor (DVS) [1], work very differently from traditional cameras (Fig. 1). They have independent pixels that send information (called "events") only in presence of brightness changes in the scene at the time they occur. Thus, their output is not an intensity image but a *stream of asynchronous events*. Event cameras excel at sensing motion, and they do so with very low-latency (1 microsecond). However, they do not provide absolute intensity measurements, rather they measure only *changes of intensity*. Conversely, standard cameras provide *direct* intensity measurements for every pixel, but with comparatively much higher latency (10–20 ms). Event cameras and standard cameras are, thus, complementary, which calls for the development of novel algorithms capable of combining the specific advantages of both cameras to perform computer vision tasks with low-latency. In fact, the Dynamic and Active-pixel Vision Sensor (DAVIS) [2] was recently introduced (2014) in that spirit. It is a sensor comprising an asynchronous event-based sensor and a standard frame-based camera in the same pixel array.

We tackle the problem of feature tracking using both events and frames, such as those provided by the DAVIS. Our goal is to combine both types of intensity measurements to maximize tracking accuracy and age, and for this reason we develop a maximum likelihood approach based on a generative event model.

Feature tracking is an important research topic in computer vision, and has been widely studied in the last decades. It is a core building block of numerous applications, such as object tracking [3] or Simultaneous Localization and Mapping (SLAM) [4–7]. While feature detection and tracking methods for frame-based cameras are well established, they cannot track in the blind time between consecutive frames, and are expensive because they process information from all pixels, even in the absence of motion in the scene. Conversely, event cameras acquire only relevant information for tracking and respond asynchronously, thus, filling the blind time between consecutive frames.

In this work we present a feature tracker which works by extracting corners in frames and subsequently tracking them using only events. This allows us to take advantage of the asynchronous, high dynamic range and low-latency nature of the events to produce feature tracks with high temporal resolution. However, this asynchronous nature means that it becomes a challenge to associate individual events coming from the same object, which is known as the data association problem. In contrast to previous works which used heuristics to solve for data association, we propose a maximum likelihood approach based on a generative event model that uses the photometric information from the frames to solve the problem. In summary, our contributions are the following:

- We introduce the first feature tracker that combines events and frames in a way that (*i*) fully exploits the strength of the brightness gradients causing the events, (*ii*) circumvents the data association problem between events and pixels of the frame, and (*iii*) leverages a generative model to explain how events are related to brightness patterns in the frames.
- We provide a comparison with state-of-the-art methods [8,9], and show that our tracker provides feature tracks that are both more accurate and longer.
- We thoroughly evaluate the proposed tracker using scenes from the publicly available Event Camera Dataset [10], and show its performance both on man-made environments with large contrast and in natural scenes.

## 2 Related Work

Feature detection and tracking with event cameras is a major research topic [8, 9, 12–18], where the goal is to unlock the capabilities of event cameras and use them to solve these classical problems in computer vision in challenging scenarios inaccessible to standard cameras, such as low-power, high-speed and high dynamic range (HDR) scenarios. Recently, extensions of popular image-based keypoint detectors, such as Harris [19] and FAST [20], have been developed for event cameras [17,18]. Detectors based on the distribution of optical flow [21] for recognition applications have also been proposed for event cameras [16]. Finally, most event-based trackers use binary feature templates, either predefined [13] or built from a set of events [9], to which they align events by means of iterative point-set–based methods, such as iterative closest point (ICP) [22].

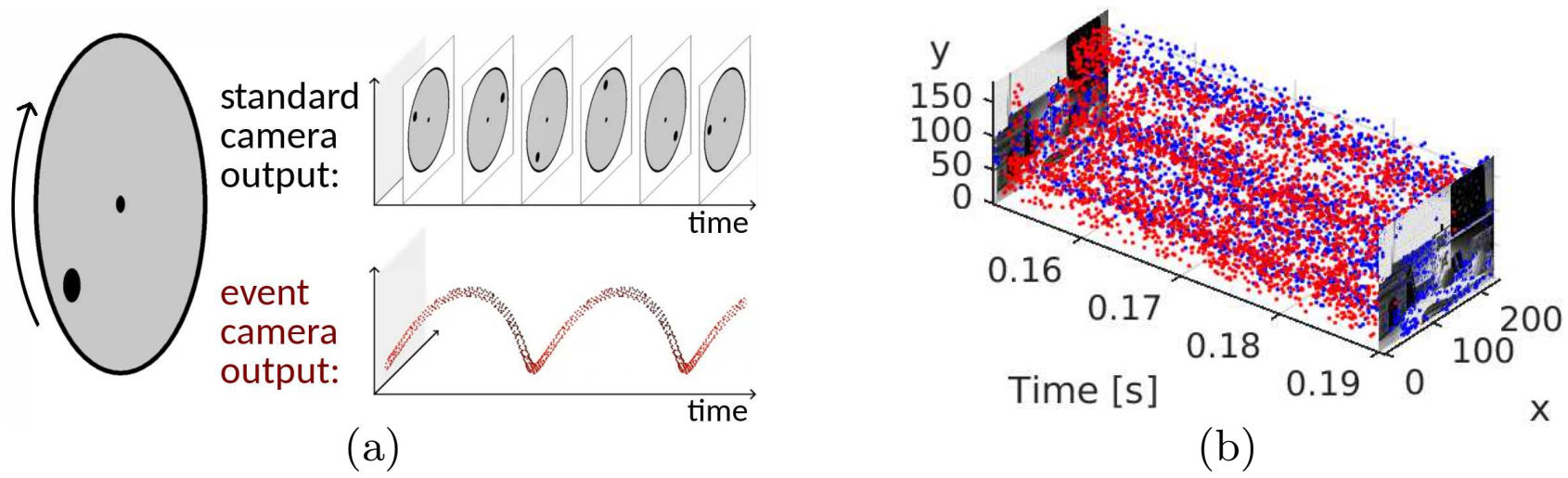

Fig. 1: Fig. 1(a): Comparison of the output of a standard frame-based camera and an event camera when facing a black dot on a rotating disk (figure adapted from [11]). The standard camera outputs frames at a fixed rate, thus sending redundant information when there is no motion in the scene. Event cameras respond to pixel-level *brightness changes* with microsecond latency. Fig. 1(b): A combined frame and event-based sensor such as the DAVIS [2] provides both standard frames and the events that occurred in between. Events are colored according to polarity: blue (brightness increase) and red (brightness decrease).

Our work is most related to [8], since both combine frames and events for feature tracking. The approach in [8] detects patches of Canny edges around Harris corners in the grayscale frames and then tracks such local edge patterns using ICP on the event stream. Thus, the patch of Canny edges acts as a template to which the events are registered to yield tracking information. Under the simplifying assumption that events are mostly generated by strong edges, the Canny edgemap template is used as a proxy for the underlying grayscale pattern that causes the events. The method in [8] converts the tracking problem into a geometric, point-set alignment problem: the event coordinates are compared against the point template given by the pixel locations of the Canny edges. Hence, pixels where no events are generated are, efficiently, not processed. However, the method has two drawbacks: ($i$) the information about the strength of the edges is lost (since the point template used for tracking is obtained from a binary edgemap) ($ii$) explicit correspondences (i.e., data association) between the events and the template need to be established for ICP-based registration. The method in [9] can be interpreted as an extension of [8] with ($i$) the Canny-edge patches replaced by motion-corrected event point sets and ($ii$) the correspondences computed in a soft manner using Expectation-Maximization (EM)-ICP.

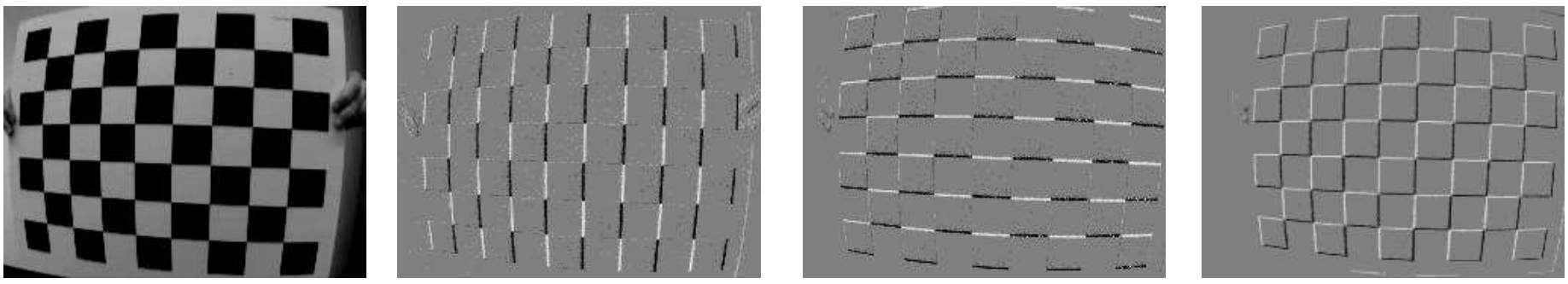

(a) Frame. (b) Left-right motion. (c) Up-down motion. (d) Diagonal motion.

Fig. 2: Result of moving a checkerboard (a) in front of an event camera in different directions. (b)-(d) show brightness increment images (Eq. (2)) obtained by accumulating events over a short time interval. Pixels that do not change intensity are represented in gray, whereas pixels that increased or decreased intensity are represented in bright and dark, respectively. Clearly, (b) (only vertical edges), (c) (only horizontal edges), and (d) cannot be related to each other without the prior knowledge of the underlying photometric information provided by (a).

Like [8, 9], our method can be used to track generic features, as opposed to constrained edge patterns. However, our method differs from [8,9] in that (*i*) we take into account the strength of the edge pattern causing the events and (*ii*) we do not need to establish correspondences between the events and the edgemap template. In contrast to [8,9], which use a point-set template for event alignment, our method uses the spatial gradient of the raw intensity image, directly, as a template. Correspondences are implicitly established as a consequence of the proposed image-based registration approach (Section 4), but before that, let us motivate why establishing correspondences is challenging with event cameras.

## 3 The Challenge of Data Association for Feature Tracking

The main challenge in tracking scene features (i.e., edge patterns) with an event camera is that, because this sensor responds to temporal changes of intensity (caused by moving edges on the image plane), the appearance of the feature varies depending on the motion, and thus, continuously changes in time (see Fig. 2). Feature tracking using events requires the establishment of correspondences between events at different times (i.e., data association), which is difficult due to the above-mentioned varying feature appearance (Fig. 2).

Instead, if additional information is available, such as the absolute intensity of the pattern to be tracked (i.e., a time-invariant representation or "map" of the feature), such as in Fig. 2(a), then event correspondences may be established indirectly, via establishing correspondences between the events and the additional map. This, however, additionally requires to continuously estimate the motion (optic flow) of the pattern. This is in fact an important component of our approach. As we show in Section 4, our method is based on a model to generate a prediction of the time-varying event-feature appearance using a given frame and an estimate of the optic flow. This generative model has not been considered in previous feature tracking methods, such as [8, 9].

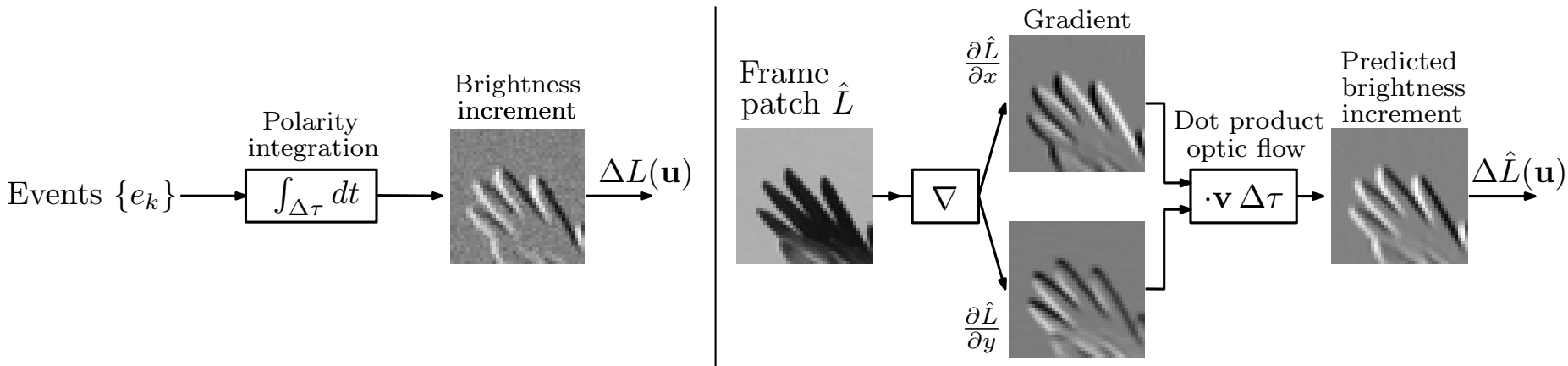

(a) $\Delta L(\mathbf{u})$ given by the events (2). (b) $\Delta \hat{L}(\mathbf{u})$ predicted using the frame (3).

Fig. 3: Brightness increments given by the events (2) vs. predicted from the frame and the optic flow using the generative model (3). Pixels of $L(\mathbf{u})$ that do not change intensity are represented in gray in $\Delta L$, whereas pixels that increased or decreased intensity are represented in bright and dark, respectively.

# 4 Methodology

An event camera has independent pixels that respond to changes in the continuous brightness signal[1] $L(\mathbf{u}, t)$. Specifically, an event $e_k = (x_k, y_k, t_k, p_k)$ is triggered at pixel $\mathbf{u}_k = (x_k, y_k)^\top$ and at time $t_k$ as soon as the brightness increment since the last event at the pixel reaches a threshold $\pm C$ (with $C > 0$):

$$\Delta L(\mathbf{u}_k, t_k) \doteq L(\mathbf{u}_k, t_k) - L(\mathbf{u}_k, t_k - \Delta t_k) = p_k C, \tag{1}$$

where $\Delta t_k$ is the time since the last event at the same pixel, $p_k \in \{-1, +1\}$ is the event polarity (i.e., the sign of the brightness change). Eq. (1) is the event generation equation of an ideal sensor [23, 24].

## 4.1 Brightness-Increment Images from Events and Frames

Pixel-wise accumulation of event polarities over a time interval $\Delta\tau$ produces an image $\Delta L(\mathbf{u})$ with the amount of brightness change that occurred during the interval (Fig. 3a),

$$\Delta L(\mathbf{u}) = \sum_{t_k \in \Delta\tau} p_k C\, \delta(\mathbf{u} - \mathbf{u}_k), \tag{2}$$

where $\delta$ is the Kronecker delta due to its discrete argument (pixels on a lattice).

For small $\Delta\tau$, such as in the example of Fig. 3a, the brightness increments (2) are due to moving edges according to the formula[2]:

$$\Delta L(\mathbf{u}) \approx -\nabla L(\mathbf{u}) \cdot \mathbf{v}(\mathbf{u}) \Delta\tau, \tag{3}$$

[1] Event cameras such as the DVS [1] respond to logarithmic brightness changes, i.e., $L \doteq \log I$, with brightness signal $I$, so that (1) represents logarithmic changes.

[2] Eq. (3) can be shown [24] by substituting the brightness constancy assumption (i.e., optical flow constraint) $\frac{\partial L}{\partial t}(\mathbf{u}(t), t) + \nabla L(\mathbf{u}(t), t) \cdot \dot{\mathbf{u}}(t) = 0$, with image-point velocity $\mathbf{v} \equiv \dot{\mathbf{u}}$, in Taylor's approximation $\Delta L(\mathbf{u}, t) \doteq L(\mathbf{u}, t) - L(\mathbf{u}, t - \Delta\tau) \approx \frac{\partial L}{\partial t}(\mathbf{u}, t)\Delta\tau$.

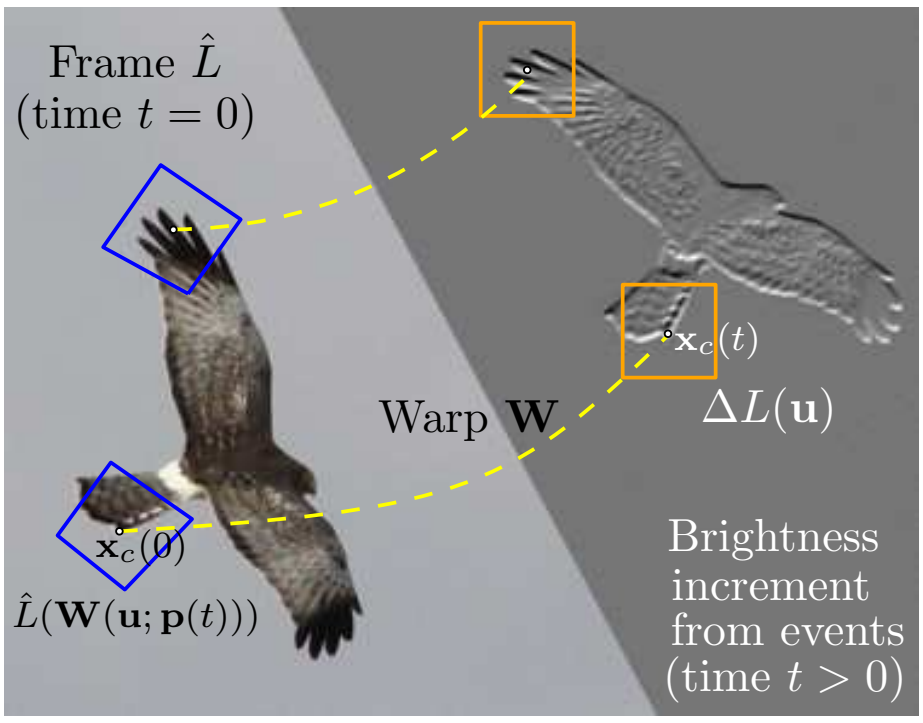

Fig. 4: Illustration of tracking for two independent patches. Events in a space-time window at time $t > 0$ are collected into a patch of brightness increments $\Delta L(\mathbf{u})$ (in orange), which is compared, via a warp (i.e., geometric transformation) $\mathbf{W}$ against a predicted brightness increment image based on $\hat{L}$ (given at $t = 0$) around the initial feature location (in blue). Patches are computed as shown in Fig. 5, and are compared in the objective function (6).

that is, increments are caused by brightness gradients $\nabla L(\mathbf{u}) = \left(\frac{\partial L}{\partial x}, \frac{\partial L}{\partial y}\right)^\top$ moving with velocity $\mathbf{v}(\mathbf{u})$ over a displacement $\Delta\mathbf{u} \doteq \mathbf{v}\Delta\tau$ (see Fig. 3b). As the dot product in (3) conveys, if the motion is parallel to the edge ($\mathbf{v} \perp \nabla L$), the increment vanishes, i.e., no events are generated. From now on (and in Fig. 3b) we denote the modeled increment (3) using a hat, $\Delta\hat{L}$, and the frame by $\hat{L}$.

### 4.2 Optimization Framework

Following a maximum likelihood approach, we propose to use the difference between the observed brightness changes $\Delta L$ from the events (2) and the predicted ones $\Delta\hat{L}$ from the brightness signal $\hat{L}$ of the frames (3) to estimate the motion parameters that best explain the events according to an optimization score.

More specifically, we pose the feature tracking problem using events and frames as that of *image registration* [25, 26], between images (2) and (3). Effectively, frames act as feature templates with respect to which events are registered. As is standard, let us assume that (2) and (3) are compared over small patches ($\mathcal{P}$) containing distinctive patterns, and further assume that the optic flow $\mathbf{v}$ is constant for all pixels in the patch (same regularization as [25]).

Letting $\hat{L}$ be given by an intensity frame at time $t = 0$ and letting $\Delta L$ be given by events in a space-time window at a later time $t$ (see Fig. 4), our goal is to find the registration parameters $\mathbf{p}$ and the velocity $\mathbf{v}$ that maximize the similarity between $\Delta L(\mathbf{u})$ and $\Delta\hat{L}(\mathbf{u};\mathbf{p},\mathbf{v}) = -\nabla\hat{L}(\mathbf{W}(\mathbf{u};\mathbf{p}))\cdot\mathbf{v}\Delta\tau$, where $\mathbf{W}$ is the warping map used for the registration. We explicitly model optic flow $\mathbf{v}$ instead of approximating it by finite differences of past registration parameters to avoid introducing approximation errors and to avoid error propagation from past noisy feature positions. A block diagram showing how both brightness increments are computed, including the effect of the warp $\mathbf{W}$, is given in Fig. 5. Assuming that the difference $\Delta L - \Delta \hat{L}$ follows a zero-mean additive Gaussian distribution with variance $\sigma^2$ [1], we define the likelihood function of the set of events $\mathcal{E} \doteq \{e_k\}_{k=1}^{N_e}$ producing $\Delta L$ as

$$p(\mathcal{E} \,|\, \mathbf{p}, \mathbf{v}, \hat{L}) = \frac{1}{\sqrt{2\pi\sigma^2}} \exp\left( -\frac{1}{2\sigma^2} \int_{\mathcal{P}} \left(\Delta L(\mathbf{u}) - \Delta \hat{L}(\mathbf{u}; \mathbf{p}, \mathbf{v})\right)^2 d\mathbf{u} \right). \quad (4)$$

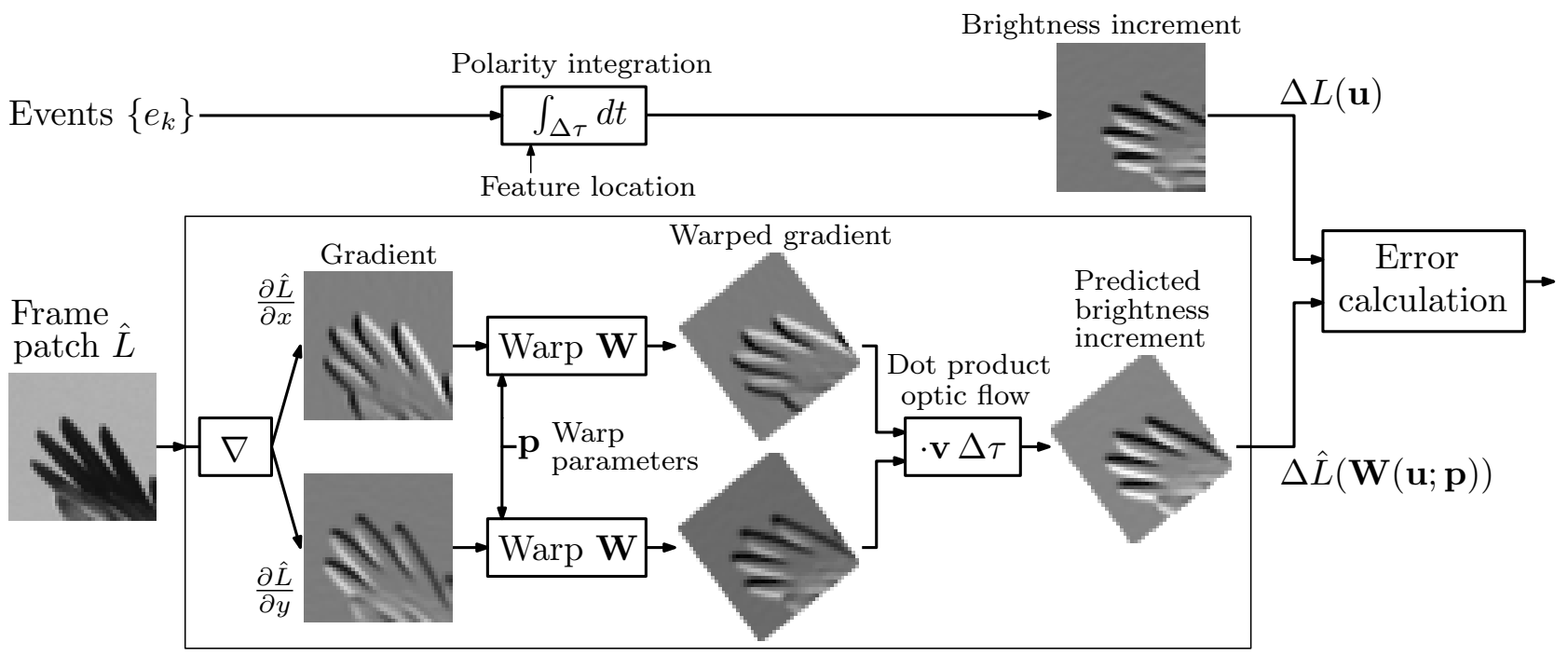

Fig. 5: Block diagram showing how the brightness increments being compared are computed for a patch of Fig. 4. Top of the diagram is the brightness increment from event integration (2). At the bottom is the generative event model from the frame (3).

Maximizing this likelihood with respect to the motion parameters $\mathbf{p}$ and $\mathbf{v}$ (since $\hat{L}$ is known) yields the minimization of the $L^2$ norm of the photometric residual,

$$\min_{\mathbf{p},\mathbf{v}} \|\Delta L(\mathbf{u}) - \Delta \hat{L}(\mathbf{u}; \mathbf{p}, \mathbf{v})\|^2_{L^2(\mathcal{P})} \quad (5)$$

where $\|f(\mathbf{u})\|^2_{L^2(\mathcal{P})} \doteq \int_{\mathcal{P}} f^2(\mathbf{u}) d\mathbf{u}$. However, the objective function (5) depends on the contrast sensitivity $C$ (via (2)), which is typically unknown in practice. Inspired by [26], we propose to minimize the difference between unit-norm patches:

$$\min_{\mathbf{p},\mathbf{v}} \left\| \frac{\Delta L(\mathbf{u})}{\|\Delta L(\mathbf{u})\|_{L^2(\mathcal{P})}} - \frac{\Delta \hat{L}(\mathbf{u}; \mathbf{p}, \mathbf{v})}{\|\Delta \hat{L}(\mathbf{u}; \mathbf{p}, \mathbf{v})\|_{L^2(\mathcal{P})}} \right\|^2_{L^2(\mathcal{P})}, \quad (6)$$

which cancels the terms in $C$ and $\Delta\tau$, and only depends on the direction of the feature velocity $\mathbf{v}$. In this generic formulation, the same type of parametric warps $\mathbf{W}$ as for image registration can be considered (projective, affine, etc.). For simplicity, we consider warps given by rigid-body motions in the image plane,

$$\mathbf{W}(\mathbf{u}; \mathbf{p}) = \mathtt{R}(\mathbf{p})\mathbf{u} + \mathbf{t}(\mathbf{p}), \quad (7)$$

where $(\mathtt{R}, \mathbf{t}) \in SE(2)$. The objective function (6) is optimized using the non-linear least squares framework provided in the Ceres software [27].

**Algorithm 1** Photometric feature tracking using events and frames

**Feature initialization:**

- Detect Harris corners [19] on the frame $\hat{L}(\mathbf{u})$, extract intensity patches around corner points and compute $\nabla\hat{L}(\mathbf{u})$.
- Set patches $\Delta L(\mathbf{u}) = 0$, set initial registration parameters $\mathbf{p}$ to those of the identity warp, and set the number of events $N_e$ to integrate on each patch.

**Feature tracking:**

**for** each incoming event **do**
  - Update the patches containing the event (i.e., accumulate polarity pixel-wise (2)).
  **for** each patch $\Delta L(\mathbf{u})$ (once $N_e$ events have been collected (2)) **do**
    - Minimize the objective function (6), to get parameters $\mathbf{p}$ and optic flow $\mathbf{v}$.
    - Update the registration parameters $\mathbf{p}$ of the feature patch (e.g., position).
    - Reset the patch ($\Delta L(\mathbf{u}) = 0$) and recompute $N_e$.

### 4.3 Discussion of the Approach

One of the most interesting characteristics of the proposed method (6) is that it is based on a generative model for the events (3). As shown in Fig. 5, the frame $\hat{L}$ is used to produce a registration template $\Delta\hat{L}$ that changes depending on $\mathbf{v}$ (weighted according to the dot product) in order to best fit the motion-dependent event data $\Delta L$, and so does our method not only estimate the warping parameters of the event-feature but also its optic flow. This optic flow dependency was not explicitly modeled in previous works, such as [8, 9]. Moreover, for the template, we use the full gradient information of the frame $\nabla\hat{L}$, as opposed to its Canny (i.e., binary-thresholded) version [8], which provides higher accuracy and the ability to track less salient patterns.

Another characteristic of our method is that it does not suffer from the problem of establishing event-to-feature correspondences, as opposed to ICP methods [8,9]. We borrow the implicit pixel-to-pixel data association typical of image registration methods by creating, from events, a convenient image representation. Hence, our method has smaller complexity (establishing data association in ICP [8] has quadratic complexity) and is more robust since it is less prone to be trapped in local minima caused by data association (as will be shown in Section 5.3). As optimization iterations progress, all event correspondences evolve jointly as a single entity according to the evolution of the warped pixel grid.

Additionally, monitoring the evolution of the minimum cost values (6) provides a sound criterion to detect feature track loss and, therefore, initialize new feature tracks (e.g., in the next frame or by acquiring a new frame on demand).

### 4.4 Algorithm

The steps of our asynchronous, low-latency feature tracker are summarized in Algorithm 1, which consists of two phases: $(i)$ initialization of the feature patch and $(ii)$ tracking the pattern in the patch using events according to (6). Multiple patches are tracked independently from one another. To compute a patch $\Delta L(\mathbf{u})$, (2), we integrate over a given number of events $N_e$ [28–31] rather than over a fixed time $\Delta\tau$ [32,33]. Hence, tracking is asynchronous, as soon as $N_e$ events are acquired on the patch (2), which typically happens at rates higher than the frame rate of the standard camera ($\sim$ 10 times higher). The supplementary material provides an analysis of the sensitivity of the method with respect to $N_e$ and a formula to compute a sensible value, to be used in Algorithm 1.

## 5 Experiments

To illustrate the high accuracy of our method, we first evaluate it on simulated data, where we can control scene depth, camera motion, and other model parameters. Then we test our method on real data, consisting of high-contrast and natural scenes, with challenging effects such as occlusions, parallax and illumination changes. Finally, we show that our tracker can operate using frames reconstructed from a set of events [34, 35], which have higher dynamic range than those of standard cameras, thus opening the door to feature tracking in high dynamic range (HDR) scenarios.

For all experiments we use patches $\Delta L(\mathbf{u})$ of $25 \times 25$ pixel size[3] and the corresponding events falling within the patches as the features moved on the image plane. On the synthetic datasets, we use the 3D scene model and camera poses to compute the ground truth feature tracks. On the real datasets, we use KLT [25] as ground truth. Since our feature tracks are produced at a higher temporal resolution than the ground truth, interpolating ground truth feature positions may lead to wrong error estimates if the feature trajectory is not linear in between samples. Therefore, we evaluate the error by comparing each ground truth sample with the feature location given by linear interpolation of the two closest feature locations in time and averaging the Euclidean distance between ground truth and the estimated positions.

### 5.1 Simulated Data. Assessing Tracking Accuracy

By using simulated data we assess the accuracy limits of our feature tracker. To this end, we used the event camera simulator presented in [10] and 3D scenes with different types of texture, objects and occlusions (Fig. 6). The tracker's accuracy can be assessed by how the average feature tracking error evolves over time (Fig. 6(c)); the smaller the error, the better. All features were initialized using the first frame and then tracked until discarded, which happened if they left the field of view or if the registration error (6) exceeded a threshold of 1.6. We define a feature's age as the time elapsed between its initialization and its disposal. The longer the features survive, the more robust the tracker.

The results for simulated datasets are given in Fig. 6 and Table 1. Our method tracks features with a very high accuracy, of about 0.4 pixel error on average, which can be regarded as a lower bound for the tracking error (under noise-free conditions). The remaining error is likely due to the linearization approximation in (3). Note that feature age is just reported for completeness, since simulation time cannot be compared to the physical time of real data (Section 5.2).

[3] A justification of the choice of patch size can be found in the supplementary material.

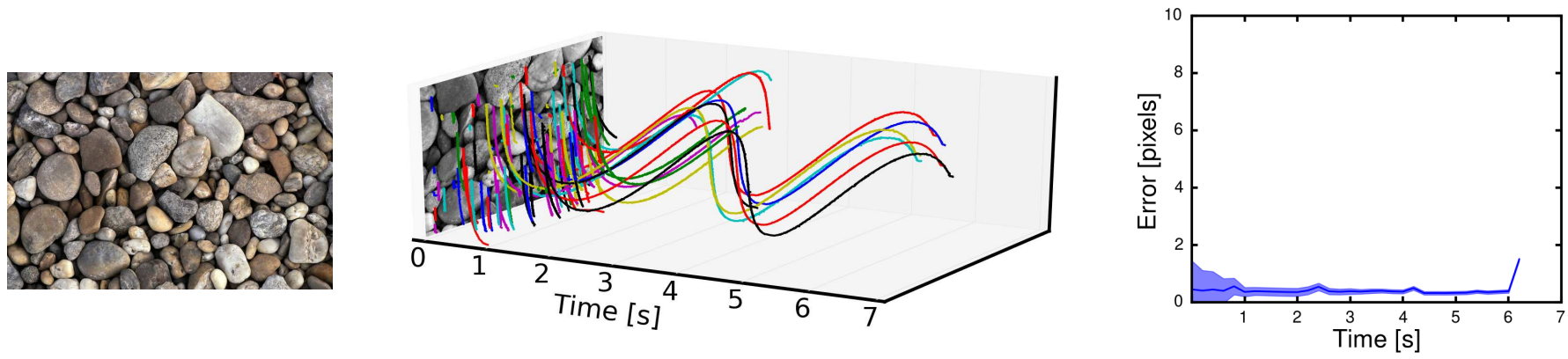

(a) (b) (c)

Fig. 6: Feature tracking results on simulated data. (a) Example texture used to generate synthetic events in the simulator [10]. (b) Qualitative feature tracks represented as curves in space-time. (c) Mean tracking error (center line) and fraction of surviving features (width of the band around the center line) as a function of time. Our features are tracked with 0.4 pixel accuracy on average.

Table 1: Average pixel error and average feature age for simulated data.

| **Datasets** | **Error** [px] | **Feature age** [s] |
|---|---|---|
| sim_april_tags | 0.20 | 1.52 |
| sim_3planes | 0.29 | 0.78 |
| sim_rocks | 0.42 | 1.00 |
| sim_3wall | 0.67 | 0.40 |

### 5.2 Real Data

We compare our method against the state-of-the-art [8, 9]. The methods were evaluated on several datasets. For [8] the same set of features extracted on frames was tracked, while for [9] features were initialized on motion-corrected event images and tracked with subsequent events. The results are reported in Fig. 7 and in Table 2. The plots in Fig. 7 show the mean tracking error as a function of time (center line). The width of the colored band indicates the proportion of features that survived up to that point in time. The width of the band decreases with time as feature tracks are gradually lost. The wider the band, the more robust the feature tracker. Our method outperforms [8] and [9] in both tracking accuracy and length of the tracks.

In simple, black and white scenes (Figs. 7(a) and 7(d)), such as those in [8], our method is, on average, twice as accurate and produces tracks that are almost three times longer than [8]. Compared to [9] our method is also more accurate and robust. For highly textured scenes (Figs. 7(b) and 7(e)), our tracker maintains the accuracy even though many events are generated everywhere in the patch, which leads to significantly high errors in [8, 9]. Although our method and [9] achieve similar feature ages, our method is more accurate. Similarly, our method performs better than [8] and is more accurate than [9] on natural scenes (Figs. 7(c) and 7(f)). For these scenes [9] exhibits the highest average feature age. However, being a purely event-based method, it suffers from drift due to changing event appearance, as is most noticeable in Fig. 7(f). Our method does not drift since it uses a time invariant template and a generative model to register events, as opposed to an event-based template [9]. Additionally, unlike previous works, our method also exploits the full range of the brightness gradients instead of using simplified, point-set–based edge maps, thus yielding higher accuracy. A more detailed comparison with [8] is further explored in Section 5.3, where we show that our objective function is better behaved.

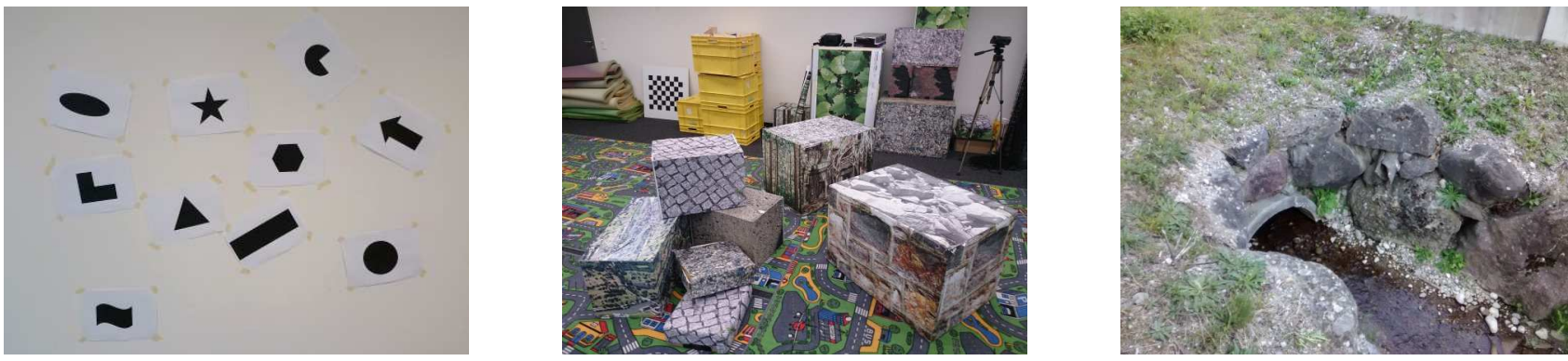

(a) Black & white scene (b) High-texture scenes (c) Natural scenes

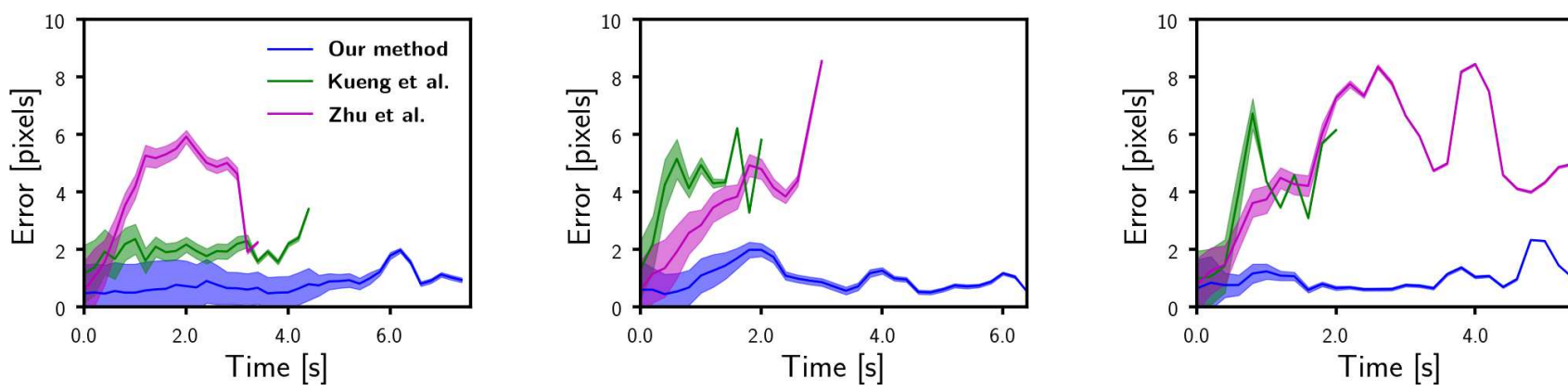

(d) shapes_6dof (e) boxes_6dof (f) rocks

Fig. 7: Feature tracking on simple black and white scenes (a), highly textured scenes (b) and natural scenes (c). Plots (d) to (f) show the mean tracking error (center line) and fraction of surviving features (band around the center line) for our method and [8, 9] on three datasets, one for each type of scene in (a)-(c). More plots are provided in the supplementary material.

Table 2: Average pixel error and average feature age for various datasets.

| Scene | Datasets | Error [px] | | | Feature age [s] | | |
|---|---|---|---|---|---|---|---|
| | | Our method | Kueng [8] | Zhu [9] | Our method | Kueng [8] | Zhu [9] |
| Black and white | shapes_6dof | **0.64** | 1.75 | 3.04 | **3.94** | 1.53 | 1.30 |
| | checkerboard | **0.78** | 1.58 | 2.36 | **8.23** | 2.76 | 7.12 |
| High Texture | poster_6dof | **0.67** | 2.86 | 2.99 | **2.65** | 0.65 | 2.56 |
| | boxes_6dof | **0.90** | 3.10 | 2.47 | **1.56** | 0.78 | 1.56 |
| Natural | bicycles | **0.75** | 3.65 | 3.66 | 1.15 | 0.49 | **1.26** |
| | rocks | **0.80** | 2.11 | 3.24 | 0.78 | 0.85 | **1.13** |

The tracking error of our method on real data is larger than that on synthetic data, which is likely due to modeling errors concerning the events, including noise and dynamic effects (such as unequal contrast thresholds for events of different polarity). Nevertheless, our tracker achieves subpixel accuracy and consistently outperforms previous methods, leading to more accurate and longer tracks.

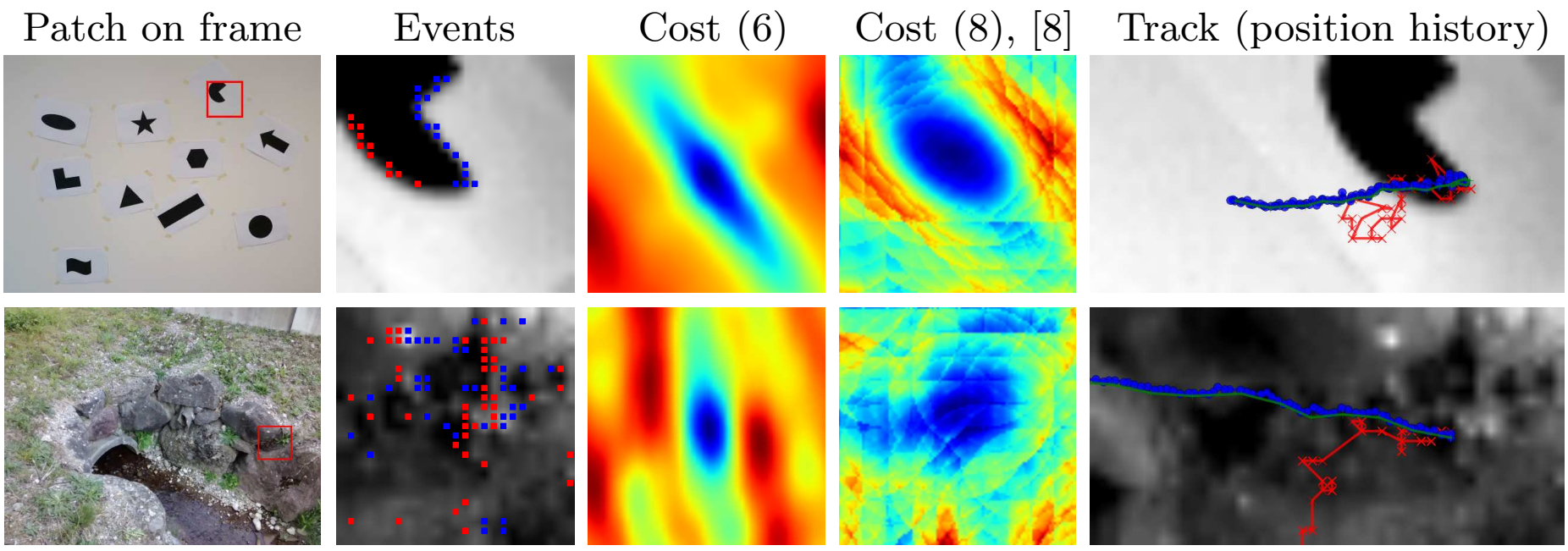

Fig. 8: Our cost function (6) is better behaved (smoother and with fewer local minima) than that in [8], yielding a better tracking (last column). The first two columns show the datasets and feature patches selected, with intensity (grayscale) and events (red and blue). The third and fourth columns compare the cost profiles of (6) and (8) for varying translation parameters in $x$ and $y$ directions ($\pm 5$ pixel around the best estimate from the tracker). The point-set–based cost used in [8] shows many local minima for more textured scenes (second row) which is not the case of our method. The last column shows the position history of the features (green is ground truth, red is [8] and blue is our method).

### 5.3 Objective Function Comparison against ICP-based Method [8]

As mentioned in Section 4, one of the advantages of our method is that data association between events and the tracked feature is implicitly established by the pixel-to-pixel correspondence of the compared patches (2) and (3). This means that we do not have to explicitly estimate it, as was done in [8, 9], which saves computational resources and prevents false associations that would yield bad tracking behavior. To illustrate this advantage, we compare the cost function profiles of our method and [8], which minimizes the alignment error (Euclidean distance) between two 2D point sets: $\{\mathbf{p}_i\}$ from the events (data) and $\{\mathbf{m}_j\}$ from the Canny edges (model),

$$\{\mathtt{R}, \mathbf{t}\} = \arg\min_{\mathtt{R},\mathbf{t}} \sum_{(\mathbf{p}_i,\mathbf{m}_i)\in \text{Matches}} b_i \left\| \mathtt{R}\mathbf{p}_i + \mathbf{t} - \mathbf{m}_i \right\|^2 . \tag{8}$$

Here, $\mathtt{R}$ and $\mathbf{t}$ are the alignment parameters and $b_i$ are weights. At each step, the association between events and model points is done by assigning each $\mathbf{p}_i$ to the closest point $\mathbf{m}_j$ and rejecting matches which are too far apart ($> 3$ pixel). By varying the parameter $\mathbf{t}$ around the estimated value while fixing $\mathtt{R}$ we obtain a slice of the cost function profile. The resulting cost function profiles for our method (6) and (8) are shown in Fig. 8.

For simple black and white scenes (first row of Fig. 8), all events generated belong to strong edges. In contrast, for more complex, highly-textured scenes (second row), events are generated more uniformly in the patch. Our method clearly shows a convex cost function in both situations. In contrast, [8] exhibits several local minima and very broad basins of attraction, making exact localization of the optimal registration parameters challenging. The broadness of the basin of attraction, together with the multitude of local minima can be explained by the fact that data association changes for each alignment parameter. This means that there are several alignment parameters which may lead to partial overlapping of the point-clouds resulting in a suboptimal solution.

To show how non-smooth cost profiles affect tracking performance, we show the feature tracks in the last column of Fig. 8. The ground truth derived from KLT is marked in green. Our tracker (in blue) is able to follow the ground truth with high accuracy. On the other hand [8] (in red) exhibits jumping behavior leading to early divergence from ground truth.

### 5.4 Tracking using Frames Reconstructed from Event Data

Recent research [34–37] has shown that events can be combined to reconstruct intensity frames that inherit the outstanding properties of event cameras (high dynamic range (HDR) and lack of motion blur). In the next experiment, we show that our tracker can be used on such reconstructed images, thus removing the limitations imposed by standard cameras. As an illustration, we focus here on demonstrating feature tracking in HDR scenes (Fig. 9). However, our method could also be used to perform feature tracking during high-speed motions by using motion-blur–free images reconstructed from events.

Standard cameras have a limited dynamic range (60 dB), which often results in under- or over-exposed areas of the sensor in scenes with a high dynamic range (Fig. 9(b)), which in turn can lead to tracking loss. Event cameras, however, have a much larger dynamic range (140 dB) (Fig. 9(b)), thus providing valuable tracking information in those problematic areas. Figs. 9(c)-(d) show qualitatively how our method can exploit HDR intensity images reconstructed from a set of events [34,35] to produce feature tracks in such difficult conditions. For example, Fig. 9(d) shows that some feature tracks were initialized in originally overexposed areas, such as the top right of the image (Fig. 9). Note that our tracker only requires a limited number of reconstructed images since features can be tracked for several seconds. This complements the computationally-demanding task of image reconstruction.

**Supplementary Material.** We encourage the reader to inspect the video, additional figures, tables and experiments provided in the supplementary material.

## 6 Discussion

While our method advances event-based feature tracking in natural scenes, there remain directions for future research. For example, the generative model we use to predict events is an approximation that does not account for severe dynamic effects and noise. In addition, our method assumes uniform optical flow in the vicinity of features. This assumption breaks down at occlusions and at objects undergoing large flow distortions, such as motion along the camera's optical axis. Nevertheless, as shown in the experiments, many features in a variety of scenes and motions do not suffer from such effects, and are therefore tracked well (with sub-pixel accuracy). Finally, we demonstrated the method using a Euclidean warp since it was more stable than more complex warping models (e.g., affine). Future research includes ways to make the method more robust to sensor noise and to use more accurate warping models.

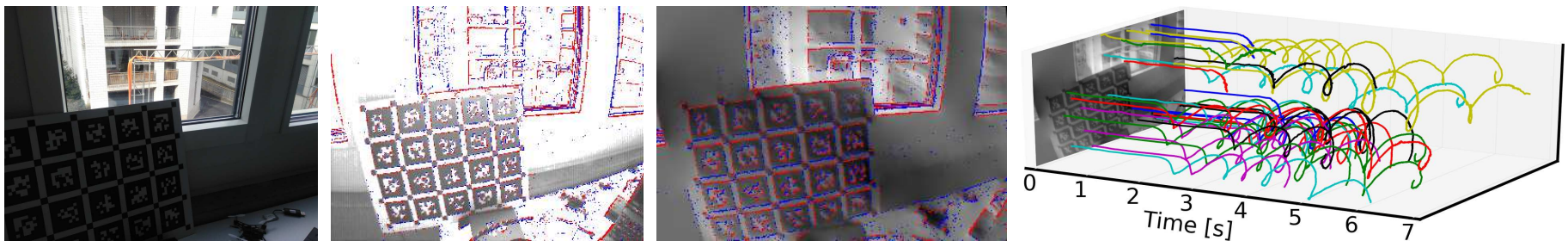

(a) (b) (c) (d)

Fig. 9: Our feature tracker is not limited to intensity frames from a real camera. In this example, we use an intensity image reconstructed from a stream of events [34,35] in a scene with high dynamic range (a). The DAVIS frame, shown in (b) with events overlaid on top, cannot capture the full dynamic range of the scene. By contrast, the reconstructed image in (c) captures the full dynamic range of the scene. Our tracker (d) can successfully use this image to produce accurate feature tracks everywhere, including the badly exposed areas of (b).

## 7 Conclusion

We presented a method that leverages the complementarity of event cameras and standard cameras to track visual features with low-latency. Our method extracts features on frames and subsequently tracks them asynchronously using events. To achieve this, we presented the first method that relates events directly to pixel intensities in frames via a generative event model. We thoroughly evaluated the method on a variety of sequences, showing that it produces feature tracks that are both more accurate (subpixel accuracy) and longer than the state of the art. We believe this work will open the door to unlock the advantages of event cameras on various computer vision tasks that rely on accurate feature tracking.

## Acknowledgment

This work was supported by the DARPA FLA program, the Swiss National Center of Competence Research Robotics, through the Swiss National Science Foundation, and the SNSF-ERC starting grant.

# 8 Supplementary Material

## 8.1 Multimedia Material

The accompanying video, https://youtu.be/A7UfeUnG6c4, shows the experiments presented in the paper in a better form than still images can convey.

## 8.2 Sensitivity with respect to the Number of Events in a Patch

As anticipated in Section 4.4 (Algorithm 1), we adaptively find the optimal number of events $N_e$ integrated in (2) to create a patch $\Delta L(\mathbf{u})$. Let us show how. As shown in (3), it is clear that $\Delta L(\mathbf{u})$ (thus $N_e$) depends on the scene texture as well as the motion. First, the larger the amount of texture (i.e., brightness gradients), the more events will be generated by the feature. Second, motion parallel to an edge prevents some events from being generated (Fig. 2).

Fig. 10 shows how the number of accumulated events $N_e$, which defines the appearance of the patch $\Delta L(\mathbf{u})$, affects the shape of the objective function (6), and, therefore, affects its minimizer. Using too few or too many events does not provide a reliable registration with respect to the frame template, either due to the fact that there is not enough information about the patch appearance conveyed by the events or because the information has been washed out by an excessive integration time. These are the left- and right-most plots in Fig. 10, respectively. Using an intermediate $N_e$ gives an event-brightness patch that captures the underlying scene texture and produces a nicely-shaped objective function with the minimizer at the correct warp and flow parameters.

We propose a simple formula to compute $N_e$ based on the the frame, $\hat{L}$, as follows. According to (2), the amount of brightness change over a patch is $\int_{\mathcal{P}} |\Delta L(\mathbf{u})| d\mathbf{u} = C\, N_e$ assuming that no events of opposite polarity are triggered at the same pixel during the short integration time $\Delta\tau$. Then, assuming that (3) is a good approximation for the event patch gives $C\, N_e \approx \int_{\mathcal{P}} |\nabla \hat{L}(\mathbf{u}) \cdot \mathbf{v} \Delta\tau| d\mathbf{u}$. Finally, considering an integration time $\Delta\tau \approx 1/\|\mathbf{v}\|$ (so that the events correspond to a displacement of the pattern of $\|\mathbf{v}\|\Delta\tau \approx 1$ pixel) and a threshold in the order of $C \approx 1$ gives

$$N_e \approx \int_{\mathcal{P}} \left| \nabla \hat{L}(\mathbf{u}) \cdot \frac{\mathbf{v}}{\|\mathbf{v}\|} \right| d\mathbf{u}. \tag{9}$$

At each time step, the newly estimated unit vector $\mathbf{v}/\|\mathbf{v}\|$ is used to compute the optimal number of events to be processed. For Fig. 10, this value is approximately the number of events in the center plot.

## 8.3 Influence of the Patch Size

As anticipated in Section 5, we provide a justification of the choice of the patch size used in our method. Tables 3, 4 and Fig. 11 report the dependency of the tracking error and the feature age with respect to the size of the patches used, from $5 \times 5$ pixels to $35 \times 35$ pixels.

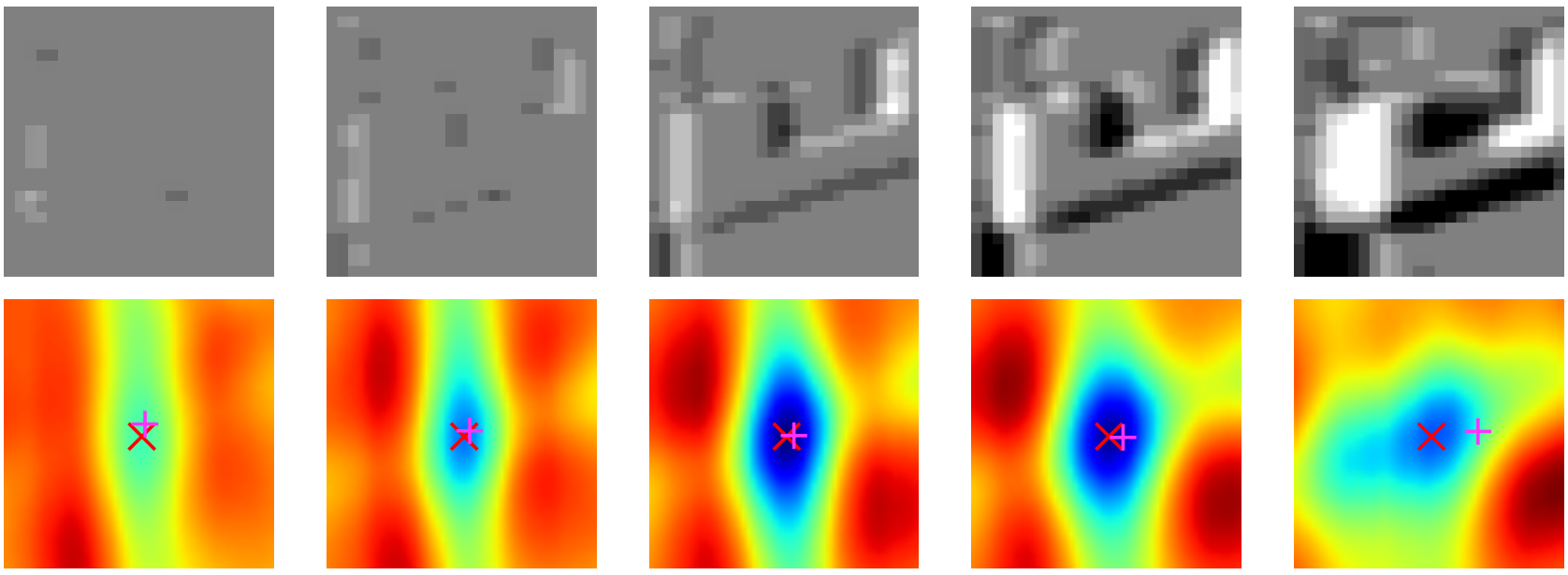

(a) $N_e = 10$ (b) $N_e = 50$ (c) $N_e = 200$ (d) $N_e = 500$ (e) $N_e = 1000$

Fig. 10: Effect of varying the number of events $N_e$ accumulated in (2). *Top row*: brightness increment patches $\Delta L(\mathbf{u})$, of size $25 \times 25$ pixels. For simplicity, the feature moves horizontally. *Bottom row*: corresponding profiles of the function (6), represented as heat maps, along the $x, y$ translation parameters ($\pm 5$ pixels from the minimizer of the function, indicated by a red cross ($\times$)). The magenta plus sign ($+$) indicates the ground truth warp parameters.

In Tables 3 and 4 we highlighted in bold the best result per row. Better accuracy is achieved for larger patch sizes whereas longer feature tracks are achieved towards medium to smaller patch sizes. We chose a patch size of $25 \times 25$ pixels as a compromise between accuracy and robustness (feature age), and performed all our experiments (Section 5) with this value.

Table 3: Tracking error for different datasets and varying patch size ($p$).

| **Datasets** | **Error [px]** | | | | | | |
|---|---|---|---|---|---|---|---|
| | $p = 5$ | $p = 11$ | $p = 15$ | $p = 21$ | $p = 25$ | $p = 31$ | $p = 35$ |
| sim_april_tags | 3.04 | 0.48 | 0.32 | 0.23 | 0.20 | 0.17 | **0.16** |
| sim_rocks | 4.61 | 1.39 | 0.55 | 0.41 | 0.42 | 0.38 | **0.35** |
| shapes_6dof | 3.62 | 0.89 | 0.65 | **0.54** | 0.64 | 0.6 | 0.62 |
| checkerboard | 2.30 | 1.25 | 1.20 | 0.93 | 0.78 | **0.75** | **0.75** |
| poster_6dof | 11.59 | 1.21 | 0.73 | 0.71 | 0.67 | **0.62** | 0.67 |
| boxes_6dof | 7.24 | 1.36 | 1.05 | 0.96 | **0.89** | 0.90 | 0.98 |
| rocks | 2.69 | 1.39 | 1.18 | 0.87 | 0.80 | 0.81 | **0.77** |
| bicycles | 3.04 | 1.20 | 1.13 | 0.88 | **0.75** | 0.83 | 0.78 |

## 8.4 Tracking Error and Feature Age. Additional Plots

Figure 12 shows the results of feature tracking on all six datasets reported in Table 2, including those that were not shown in Fig. 7 of the main text due to space limitations (Section 5). The three columns indicate the three types of scenes considered: black and white, high-texture and natural scenes. The longest tracks are achieved for black and white scenes (e.g., 25 s for the checkerboard dataset). On average, high-contrast and high-texture scenes yield longer tracks than natural scenes (width of the band around the center line). In many cases, the average error decreases as time progresses since the features that survive longest are typically those that are most accurately tracked.

Table 4: Feature age for different datasets and varying patch size ($p$).

| Datasets | Feature Age [s] | | | | | | |
|---|---|---|---|---|---|---|---|
| | $p = 5$ | $p = 11$ | $p = 15$ | $p = 21$ | $p = 25$ | $p = 31$ | $p = 35$ |
| sim_april_tags | 0.23 | 0.98 | **2.33** | 1.93 | 1.52 | 1.44 | 1.20 |
| sim_rocks | 0.05 | **1.05** | 0.72 | 0.99 | 1.00 | 0.74 | 0.86 |
| shapes_6dof | 0.59 | 3.15 | 3.31 | 3.52 | **3.97** | 3.11 | 3.21 |
| checkerboard | 2.68 | 7.72 | 8.21 | **8.32** | 8.24 | 7.74 | 8.22 |
| poster_6dof | 0.46 | 1.88 | 2.34 | 2.09 | **2.65** | 1.73 | 1.62 |
| boxes_6dof | 0.50 | 1.77 | 1.76 | **1.95** | 1.56 | 1.71 | 1.81 |
| rocks | 0.72 | 1.05 | 0.77 | 1.45 | 0.78 | **1.62** | 1.04 |
| bicycles | 0.44 | 1.22 | **1.33** | 1.26 | 1.16 | 1.19 | 1.11 |

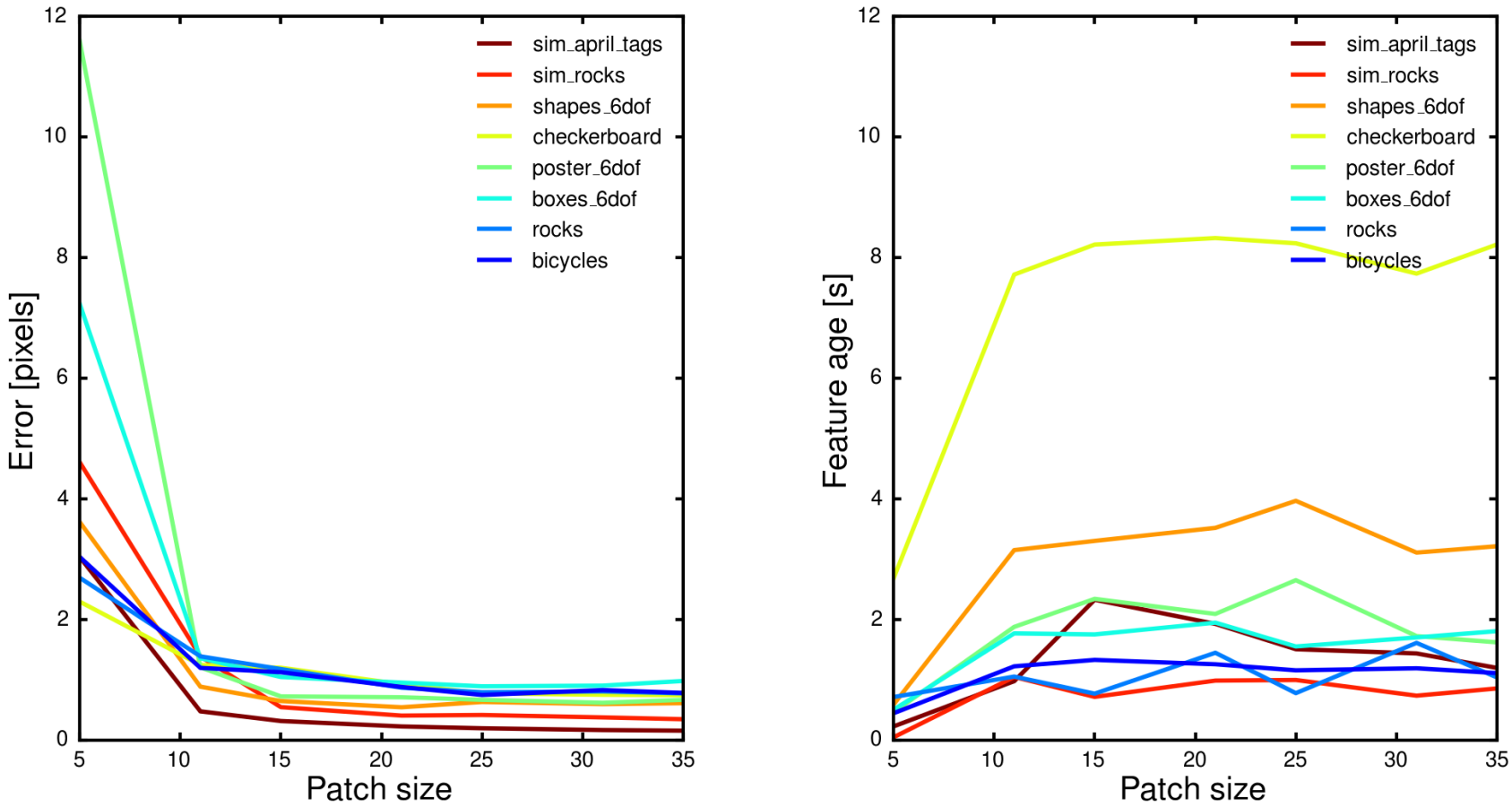

(a) (b)

Fig. 11: Fig. (a) and (b) show, respectively, the evolution of the mean tracking error and feature age as a function of the patch size used. This is a visualization of the values on Tables 3 and 4.

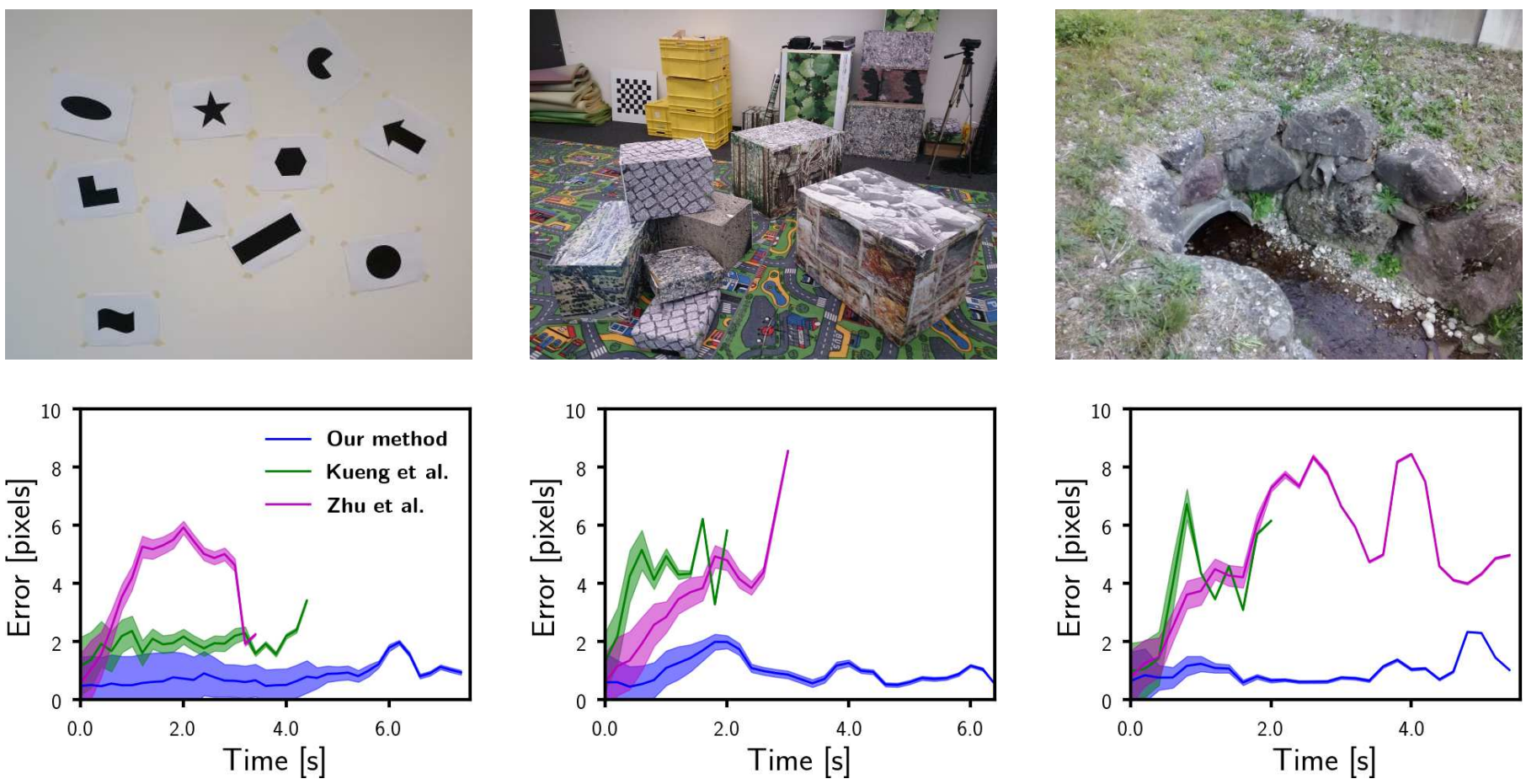

(d) shapes_6dof (e) boxes_6dof (f) rocks

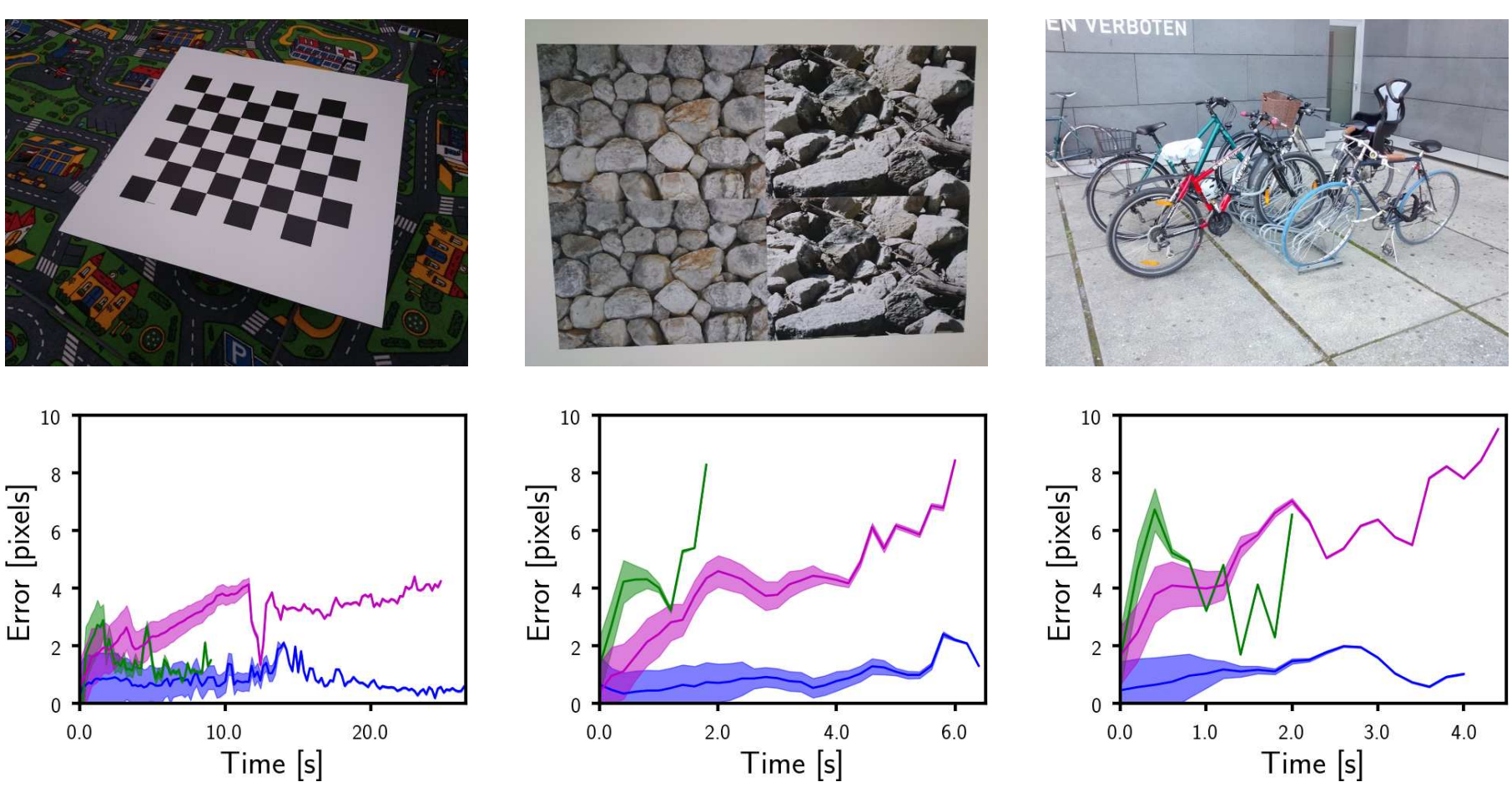

(j) checkerboard (k) poster_6dof (l) bicycles

Fig. 12: Feature tracking on all six datasets reported in Table 2. Plots (d) to (f) and (j) to (l) show the mean tracking error (center line) and fraction of surviving features (band around the center line) for our method and [8, 9].

## 8.5 Feature Tracking in Low-Light and with Abrupt Light Changes

To further illustrate the robustness of our tracker, we performed additional experiments in low-light and with abrupt changes of illumination, achieved by switching the lights on and off in the room. Results are displayed in Figs. 13, 14 and 15. In these experiments we show that our tracker can extract features from a standard frame and track them robustly through time, even when the light is off, thanks to the very high dynamic range of the event camera. Our method is also able to track after the light has been switched on again. By contrast, KLT [25] loses track immediately after switching the light off because the frames do not have as high dynamic range as the events. We encourage the reader to watch the attached video, which shows the experiment in a better form than still images can convey.

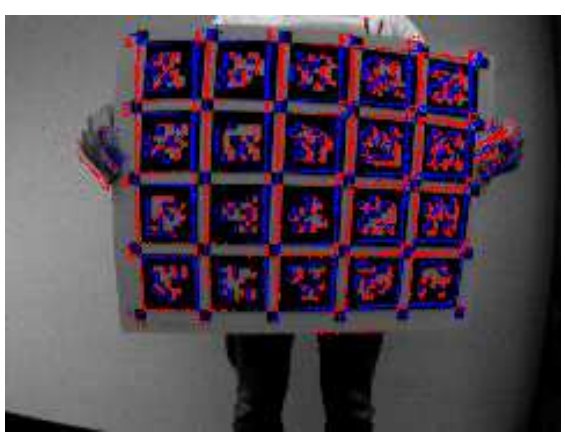

(a) Frame at $t = 0$

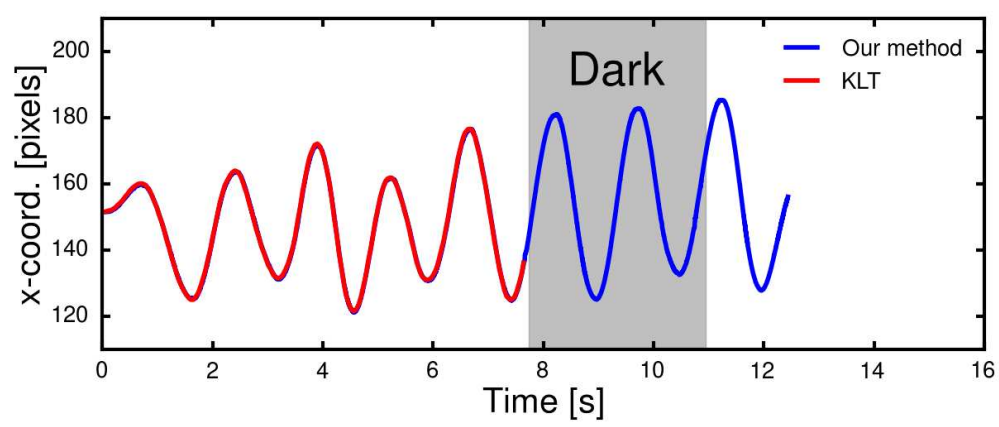

(b)

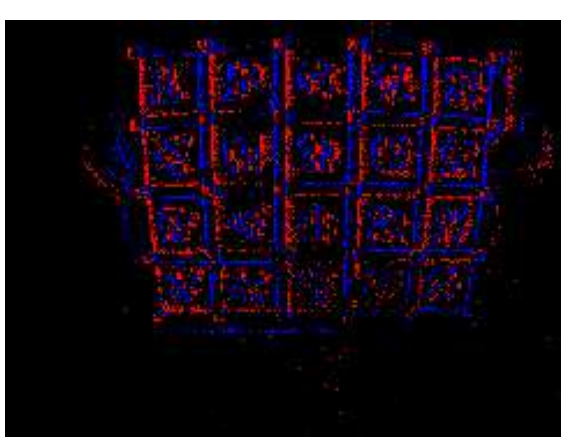

(c) Frame at time $t = 9\,\mathrm{s}$

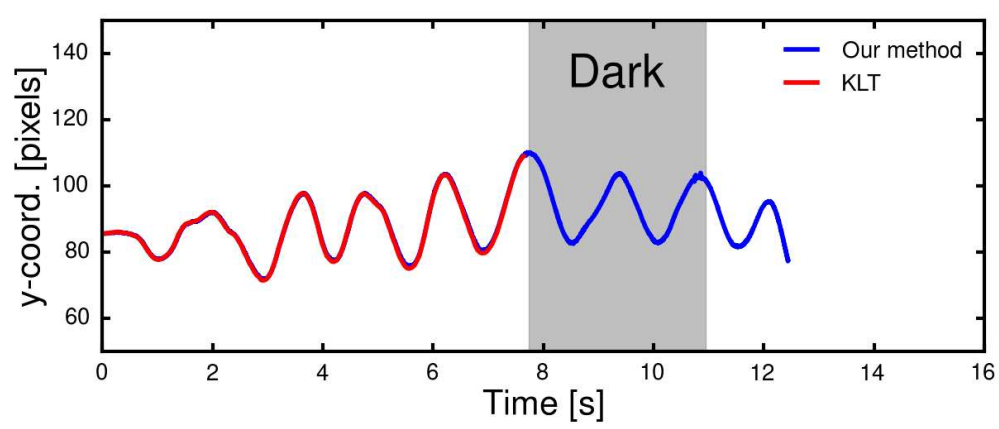

(d)

Fig. 13: Figs. (a) and (b) show the standard frames with the events superimposed, respectively when the light in the room is on or off. Figs. (b) and (d) show the evolution of the $x$ and $y$ coordinates of one feature tracked through time (red: KLT [25] on the frames, blue: our method). In contrast to KLT, our tracker maintains stable feature tracks even in the period when the light is off (marked in gray), and keeps tracking them when the light is on again.

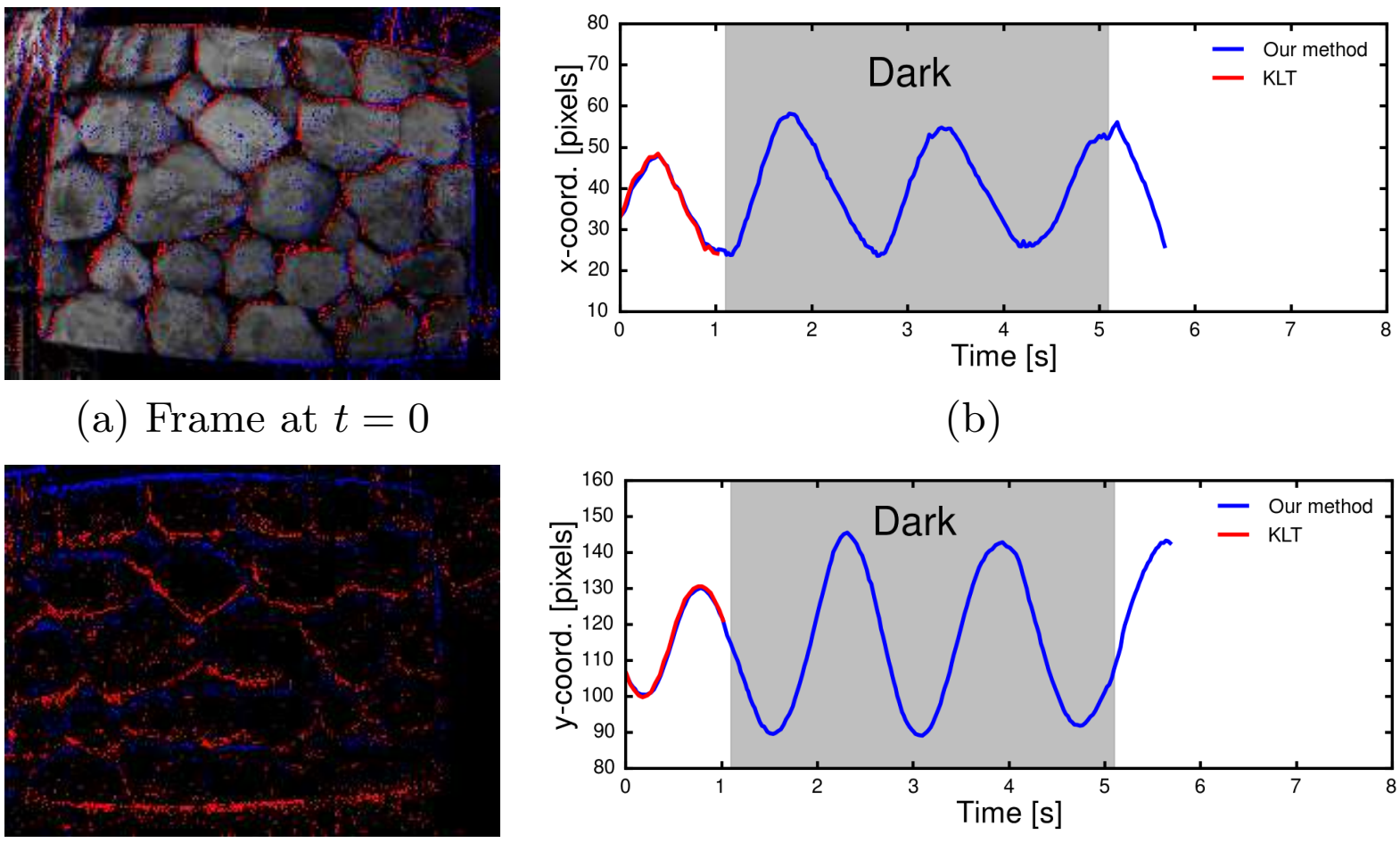

(a) Frame at $t = 0$ (b)

(c) Frame at time $t = 2\,\mathrm{s}$ (d)

Fig. 14: Feature tracking in low-light and with abrupt illumination changes. Rocks scene. Same notation as in Fig. 13

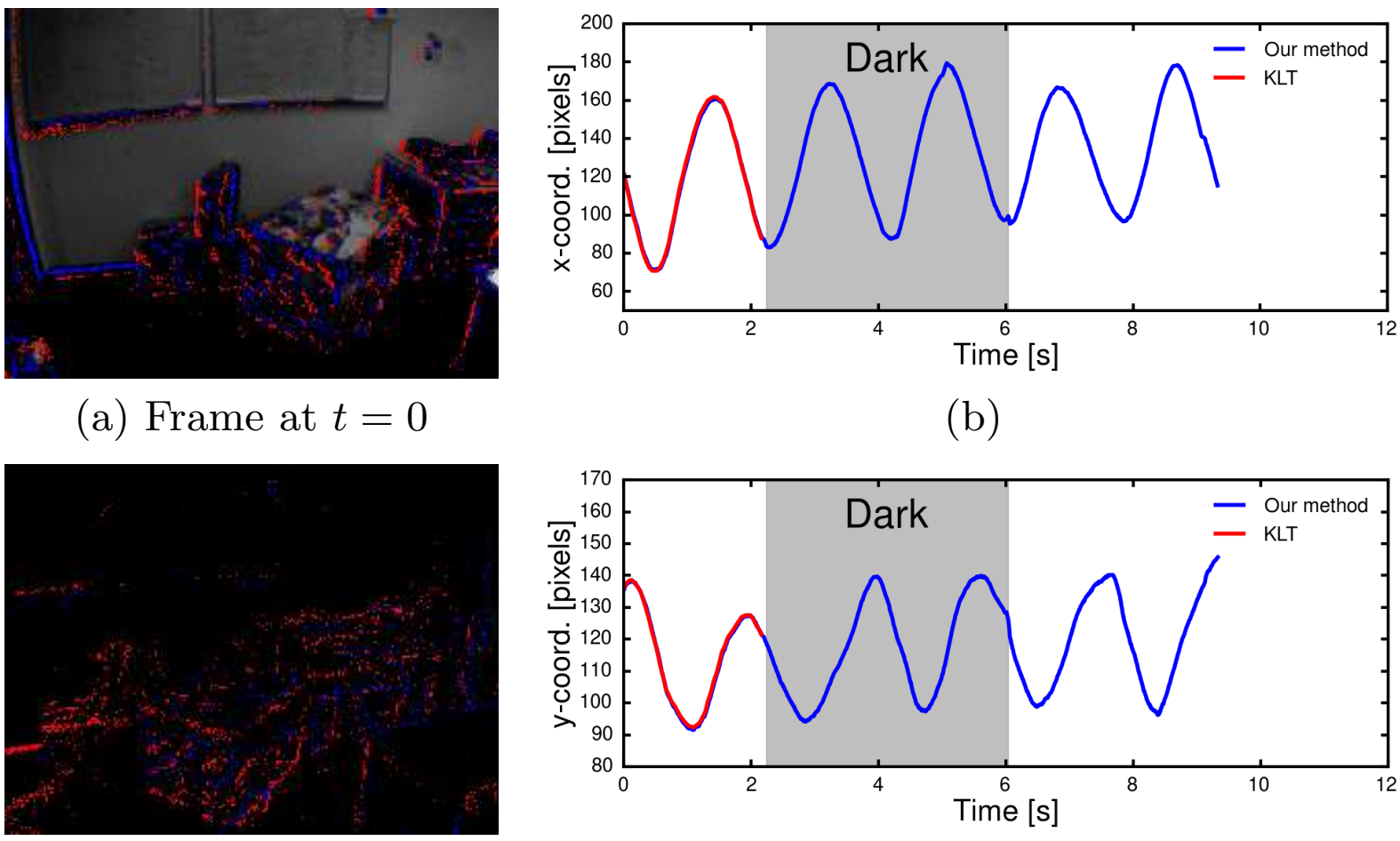

(a) Frame at $t = 0$ (b)

(c) Frame at time $t = 3\,\mathrm{s}$ (d)

Fig. 15: Feature tracking in low-light and with abrupt illumination changes. Office scene. Same notation as in Fig. 13